\documentclass{article}

\usepackage{arxiv}
\usepackage[utf8]{inputenc} 
\usepackage[T1]{fontenc}    
\usepackage{hyperref}       
\usepackage{url}            
\usepackage{booktabs}       
\usepackage{amsfonts}       
\usepackage{nicefrac}       
\usepackage{microtype}      
\usepackage{lipsum}		
\usepackage{graphicx}
\usepackage{doi}

\usepackage{microtype}%
\usepackage{relsize,comment}
\usepackage{cite}
\usepackage{pgfplots}
\pgfplotsset{compat=1.17}
\usepackage{float}
\usepackage{tikz}
\usetikzlibrary{arrows}
\newcommand{\TD}{\mathrm{TD}}
\newcommand{\TDim}{\mathrm{TDim}}
\usepackage{amsmath, amsthm, amssymb}

\newtheorem{proposition}{Proposition}
\usepackage{algorithm}
\usepackage{algpseudocode}

\title{ Exact Learning of Qualitative Constraint Networks from Membership Queries}

\author{
	\href{https://orcid.org/0000-0001-7381-1064}{\includegraphics[scale=0.06]{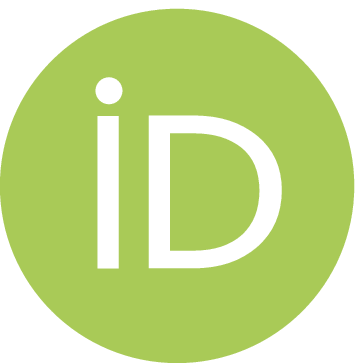}\hspace{1mm}Malek Mouhoub} \\
 	Department of Computer Science\\
	  University of Regina\\
	\texttt{mouhoubm@uregina.ca} \\
	\AND
	Hamad Al Marri \\
		Department of Computer Science\\
	  University of Regina\\
	\texttt{almarrih@uregina.ca} \\
	\And
   Eisa Alanazi\\
   Um Al Qura University Aif Road, 21955\\
Mecca, Makkah Province, Saudi Arabia \\
	\texttt{eaanazi@uqu.edu.sa} \\
}

\renewcommand{\shorttitle}{ Exact Learning of QCNs from Membership Queries}

\hypersetup{
pdftitle={Exact Learning of Qualitative Constraint Networks from Membership Queries},
pdfsubject={ },
pdfauthor={ Malek Mouhoub, Hamad Al Marri and Eisa Alanazi},
pdfkeywords={Qualitative Constraint Network (QCN), Constraint Acquisition; Exact Learning, Temporal and Spatial Reasoning, Membership Query, Path Consistency, Constraint Propagation},
}

\begin{document}
\maketitle

\begin{abstract}
 A Qualitative Constraint Network (QCN) is a constraint graph for representing problems under qualitative temporal and spatial relations, among others. More formally, a QCN includes a set of entities, and a list of qualitative constraints defining the possible scenarios between these entities. These latter constraints are expressed as disjunctions of binary relations capturing the (incomplete) knowledge between the involved entities. QCNs are very effective in representing a wide variety of real-world applications, including scheduling and  planning, configuration and Geographic Information Systems (GIS).
It is however challenging to elicit, from the user, the QCN representing a given problem. To overcome this difficulty in practice, we propose a new algorithm for learning, through membership queries, a QCN from a non expert. In this paper,  membership queries are asked in order to elicit temporal or spatial relationships between pairs of temporal or spatial entities. In order to improve the time performance of our learning algorithm in practice, constraint propagation, through transitive closure, as well as ordering heuristics, are enforced. The goal here is to reduce the number of membership queries needed to reach the target QCN.  In order to assess the practical effect of constraint propagation and ordering heuristics, we conducted several experiments  on randomly generated temporal and spatial constraint network instances. The results of the experiments are very encouraging and promising.

\end{abstract}


\section{Introduction}
\label{sec:introduction}

Temporal and spatial reasoning is very relevant in many real-world applications including scheduling, planning, Geographic Information Systems (GIS) and computational linguistics. Handling the qualitative aspects of time and space is indeed crucial when designing the related systems especially when dealing with incomplete information. Several research works have therefore been proposed in order to represent and reason on (incomplete) symbolic temporal and spatial information. One of the most known approaches is the Allen Algebra \cite{allen1983maintaining}, based on the notion of time intervals and their binary relations. Here, a time interval $I$ is an ordered pair $(I^{-},I^{+})$ such that $I^{-} < I^{+}$, where $I^{-}$ and $I^{+}$ are points on the time line. There are thirteen basic relations (Allen primitives) that can hold between intervals. A binary relation between two intervals is represented by the disjunction of some Allen primitives and expresses the incompleteness of the related temporal information. A given problem under temporal constraints can be converted into an Interval Algebra (IA) network (also called qualitative temporal  network) where nodes correspond to intervals and each arc represents the binary relation between the corresponding intervals. Note that an IA network is a particular case of a Qualitative Constraint Networks (QCN) which is a general network for representing problems under qualitative temporal or spatial constraints. Given an IA network, and more generally a QCN, the main reasoning task is to decide its consistency and return one or more solutions (consistent scenarios satisfying all the qualitative constraints) if it is the case. These tasks can be achieved with a  backtrack search algorithm enhanced with constraint propagation techniques as well as variable ordering heuristics.

QCNs have proven to be effective in representing a wide variety of temporal and spatial applications. In this regard, the main challenging task is the representation of these problems as a QCN. Indeed, this latter requires a good experience and expertise in temporal and spatial representation and reasoning. To overcome this difficulty, we propose a new algorithm for learning, through membership queries, a QCN  network from a non expert. More precisely,  membership queries are asked in order to elicit temporal or spatial relationships between pairs of temporal or spatial entities. During this acquisition process, constraint propagation, through transitive closure, as well as ordering heuristics are performed in order to reduce the number of membership queries needed to reach the target QCN.

In order to assess the time performance of our algorithm, especially when enforcing the two techniques we listed,  several experiments have been conducted on randomly generated QCN instances (considering temporal or spatial constraint networks). In all the experiments, variants of our proposed algorithm have been compared, in terms of time cost, to the well known CONACQ.2 constraint acquisition method \cite{bessiere2015constraint}. The results of the experiments clearly demonstrate the superiority of our algorithm, as well as the effect of transitive closure and some of the ordering heuristics to reduce the number queries needed to learn the target QCN.

While constraint learning (also called constraint acquisition), through membership queries, has already been reported in the literature \cite{bessiere2007query,bessiere2015constraint,bessiere2013constraint,arcangioli2015,tsouros2018efficient}, the originality of our work can be highlighted as follows.
\begin{enumerate}
\item No work, from past literature on constraint acquisition, has addressed the particular case of temporal and spatial constraints, or Qualitative Constraint Networks (QCNs) in general.
\item While the known constraint acquisition algorithms can be adapted to learn QCNs, they are time expensive especially when requiring a large number of queries to learn a given problem, which makes them impractical. In contrast, our algorithm takes advantage of constraint propagation (enforced by transitive closure) and queries ordering heuristics (following the fail-first principle) in order to dramatically reduce the required number of membership queries for learning the constraint problem. This makes our algorithm significantly superior in time efficiency, in practice, than these constraint learning algorithms, as per the experimental results we reported in Section \ref{sec:Experimentation}.
\item The scenarios learnt by our proposed algorithm (QCNs in case 1 and case 2) are guaranteed to be consistent. This is not the case of CONACQ.2, and other constraint acquisition algorithms in the literature, which assume that the constraint network elicited from the user is consistent. For QCNs in case 3, our algorithm learns path consistent constraint networks. This makes the solving part of the problem easier (as we are dealing with a smaller search space). CONACQ.2 and other learning algorithms do not have this feature, and the target constraint network obtained by these methods can be inconsistent.
\item  Constraint acquisition algorithms assume that user's answers are correct, which is not always the case in practice. If some of the answers are inconsistent, then these learning algorithms will simply fail to provide the target constraint network. In this regard, we propose a variant of our learning algorithm that is capable to identify and correct wrong user's answer(s) causing the learning process to collapse. More precisely, using transitive closure and a variant of the backtrack search method, the algorithm we propose has the ability to detect the source of inconsistency. It will then ask the user to correct and confirm the faulty answers. The new answers are again checked and the process continues until the algorithm converges to the consistent target QCN.   To our best knowledge there is no algorithm from the literature that is capable of learning with mistakes.

\item We present a study of the sample complexity for learning QCNs, in each of the three cases we consider in this paper (Section 5). In this regard, the bounds for the teaching dimension complexity parameter have been defined. To our best knowledge, this is the first time such analysis is reported in the literature.

\end{enumerate}

This paper is an extension of \cite{DBLP:conf/time/MouhoubMA18}, which focused on learning temporal constraints. We are considering in this work general QCNs including the spatial case. Moreover, many experiments reported in this paper are compared to CONACQ.2 algorithm \cite{bessiere2015constraint} (which is not the case of the conference paper).

The rest of the paper is structured as follows. Sections \ref{sec:QCN_AND_TCN} and \ref{sec:ConstraintAcquisition} respectively introduce the background knowledge on  QCNs and exact learning algorithms, based on membership queries.
 Section \ref{sec:ProposedAlgorithm} presents our proposed algorithm for learning QCNs. In section \ref{sec:TheoreticalLimits}, we list the theoretical limits for learning QCNs. The experiments we conducted to assess the time performance of our learning algorithm are reported in Section  \ref{sec:Experimentation}. Finally, concluding remarks and ideas for future works are listed in Section \ref{sec:ConclusionAndFutureWork}.


\section{Qualitative Constraint Networks (QCNs)}
\label{sec:QCN_AND_TCN}


\subsection{QCN Definition}

A Qualitative Constraint Network (QCN) is a pair $(V,C)$ in which $V$ is a finite set of variables representing temporal or spacial entities and $C$ is a finite set of constraints on these variables. Each constraint $c_i$ is expressed as a disjunction of binary relations between the involved variables. Each of these relations is defined on a language set
$\mathcal{B}=\{b_1,b_2,\dots,b_p\}$ where $p>0$. A particular case of such networks is the  IA network \cite{allen1983maintaining} where $V$ is a set of temporal events, each representing an assertion over a time interval; and $\mathcal{B}$ is the set of Allen primitives depicted in Table \ref{allentable} \footnote{Note that, six of the seven primitives listed have inverse relations. This will bring the total of number of possible relations to 13. Inverse relations can also be denoted by adding the following symbol (instead of the letter ``i''): ``$\smile$'' as in $D^{\smile}$, expressing the inverse of the relation $D$. In this  paper, we use both notations interchangeably.}. For instance, let us consider two temporal events, $E_1$ and $E_2$. The following constraint expresses the fact that both events are mutually exclusive: $ E_1\;   (B \vee  Bi)\; E_2.$ Note that a universal relation, corresponding to the disjunction of all relations within $\mathcal{B}$ ($I$, representing the 13 Allen primitives in the case of the Interval Algebra), is used to express the fact that there is no constraint between the involved entities (the knowledge on such constraint is completely unknown).

{\relsize{-1}
\begin{table}
\caption{Allen's primitives}
\begin{center}
\begin{tabular}{|l|c|c|l|l|}\hline
Relation      & Symbol & Inverse    & Meaning & Endpoints \\  \hline
&&&&\\
X precedes Y     & $P$     & $Pi$              & XXX YYY & $X^{+} < Y^{-}$ \\ \hline
&&&&\\
X equals  Y      & $E$     & $E $               & XXX & $X^{-} = Y^{-}$\\
              &         &                   & YYY  &  $X^{+} = Y^{+}$\\ \hline
&&&&\\
X meets Y & $M$     & $Mi$              & XXXYYY & $X^{+} = Y^{-}$\\ \hline
&&&&\\
X overlaps Y  & $O$     & $Oi$              & XXXX & $X^{-} < Y^{-} < X^{+}$\\
              &         &                   & ~~~~~YYYY & \\ \hline
&&&&\\
X during y   & $D$     & $Di$              & ~~~XXX & $X^{-} < Y^{-}$ \\
              &         &                   & YYYYYY &\\ \hline
&&&&\\
X starts Y    & $S$     & $Si$              & XXX & $X^{-} = Y^{-}$\\
              &         &                   & YYYYY &  \\ \hline
&&&&\\
X finishes Y   & $F$     & $Fi$              & ~~~~\,XXX & $X^{-} < Y^{-}$ \\
              &         &                   & YYYYY & \\ \hline
\end{tabular}
\end{center}

\label{allentable}
\end{table}
}

In order to illustrate the representation of a given temporal problem using an IA network, let us consider the following example \cite{Belouaer12}.

\noindent
\subsubsection*{Example\,1\,: Soccer Game}

\begin{quote}
{\em
\begin{enumerate}
\item John, Mary and Wendy {\bf separately} rode to the soccer game.
\item John either {\bf started or arrived} just as Mary {\bf started}.
\item John's trip {\bf overlapped} the soccer game.
\item Mary's trip took place {\bf during} the game or else the game took
  place {\bf during} her trip.
\end{enumerate}
}
\end{quote}

\begin{figure}
    \centerline{\resizebox{5cm}{!}{\includegraphics{{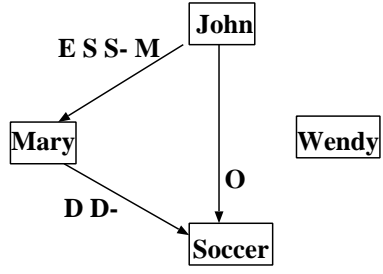}}}}
 \caption{AI network representing the Soccer Game problem \cite{DBLP:conf/appinf/Mouhoub03}}
  \label{tcsp}
\end{figure}

The above story includes four main events\,:
 John, Mary and Wendy are going to the soccer game respectively and the
 soccer game itself. The qualitative temporal relations between these events are highlighted in bold. The IA network expressing this story is depicted in Figure \ref{tcsp}. Here, nodes represent the four events and arcs are labeled with the IA relations (disjunctions of Allen primitives). For instance, the relation $E \vee S \vee S^{\smile} \vee M$ represents the temporal information listed in the second item above.

\subsection{Path Consistency for QCNs}

Given a QCN, the main reasoning task that can be performed consists of checking the consistency of the network and, if it is the case, returns one or more consistent scenarios satisfying all the constraints of the related problem. This task can be done using a backtrack search algorithm. Constraint propagation, before and during search, can be enforced in order to reduce the size of the search space by removing some of the inconsistent relations. Consequently, this will prevent late failure earlier. The most known constraint propagation technique used in QCN networks in general, and IA networks in particular, is Path Consistency (PC) \cite{mohr1986arc,van1996design,bliek1999path,long2016efficient}. This technique (also called transitive closure) applies local consistency on every subset of three variables which results in removing some inconsistent relations from the network. This will lead to the reduction on the search space. Assuming $X$, $Y$ and $Z$ are three temporal events, if $X$ precedes $Y$ and $Y$ meets $Z$, then it must be that $X$ precedes $Z$ as well. Any other relation belonging to the constraint between $X$ and $Z$ should be removed. Local consistency is enforced using a $|\mathcal{B}|^2$ composition table. Table \ref{compIA} lists the 13 x 13 compositions in the case of Allen's primitives.  Note that PC can also infer new relations by refining universal relations into a more specific one. For instance, if we take the previous example and assume that there is no constraint between $Y$ and $Z$ then PC will infer a new relation (precedes) between these two events.


\begin{table*}
\caption{Allen's composition table \cite{DBLP:journals/air/Mouhoub04}}
\label{compIA}
\begin{center}
{\relsize{-2}
\begin{tabular}{|l||*{13}{c|}}\hline
 &E&$P$&$P^{\smile}$&D&$D^{\smile}$&$O$&$O ^{\smile}$&M&$M^{\smile}$&S&$S^{\smile}$&$F$&$F^{\smile}$\\ \hline \hline
E&E&$P$&$P^{\smile}$&$D$&$D^{\smile}$&$O$&$O ^{\smile}$&$M$&m&S&s&$F$&$F^{\smile}$ \\ \hline
$P$&$P$&$P$&$I$&$u$&$P$&$P$&$u$&$P$&$u$&$P$&$P$&$u$&$P$ \\ \hline
$P^{\smile}$&p&$I$&$P^{\smile}$&$v^{\smile}$&$P^{\smile}$&$v^{\smile}$&$P^{\smile}$&$v^{\smile}$&$P^{\smile}$&$v^{\smile}$&$P^{\smile}$&$P^{\smile}$&$P^{\smile}$ \\ \hline
$D$&$D$&P&$P^{\smile}$&$D$&$I$&$u$&$v^{\smile}$&$P$&$P^{\smile}$&$D$&$v^{\smile}$&$D$&$u$ \\ \hline
$D^{\smile}$&$D^{\smile}$&$v$&$u ^{\smile}$&$n$&$D^{\smile}$&$z^{\smile}$&$y^{\smile}$&$z^{\smile}$&$y^{\smile}$&$z^{\smile}$&$D^{\smile}$&$y^{\smile}$&$D^{\smile}$ \\ \hline
$O$&$O$&$P$&$u ^{\smile}$&$y$&$v$&$x$&$n$&$P$&$y^{\smile}$&$O$&$z^{\smile}$&$y$&$x$ \\ \hline
$O ^{\smile}$&$O ^{\smile}$&$v$&$P^{\smile}$&$z$&$u ^{\smile}$&$n$&$x^{\smile}$&$z^{\smile}$&$P^{\smile}$&$z$&$x^{\smile}$&$O ^{\smile}$&$y^{\smile}$ \\ \hline
M&$M$&$P$&$u ^{\smile}$&$y$&$P$&$P$&$y$&$P$&$a$&$M$&$M$&$y$&$P$ \\ \hline
$M^{\smile}$&$M^{\smile}$&$v$&$P^{\smile}$&$z$&$P^{\smile}$&$z$&$P^{\smile}$&$b$&$P^{\smile}$&$z$&$P^{\smile}$&$M^{\smile}$&$M^{\smile}$ \\ \hline
S&S&P&$P^{\smile}$&$D$&$v$&$x$&$z$&P&$M^{\smile}$&S&$b$&$D$&$x$ \\ \hline
$S^{\smile}$&s&$v$&$P^{\smile}$&$z$&$D^{\smile}$&$z^{\smile}$&$O ^{\smile}$&$z^{\smile}$&m&$b$&s&$O ^{\smile}$&$D^{\smile}$ \\ \hline
$F$&$F$&P&$P^{\smile}$&$D$&$u ^{\smile}$&$y$&$x^{\smile}$&M&$P^{\smile}$&$D$&$x^{\smile}$&$F$&$a$ \\ \hline
$F^{\smile}$&$F^{\smile}$&P&$u
^{\smile}$&$y$&$D^{\smile}$&$O$&$y^{\smile}$&$M$&$y^{\smile}$&$O$&$D^{\smile}$&$a$&$F^{\smile}$
\\ \hline
\end{tabular}
}
\end{center}

\noindent
$x=P \vee O \vee M$\\
$y=D \vee O \vee S$\\
$z=D \vee O ^{\smile} \vee F$ \\
$a= E \vee F \vee  F^{\smile}$ \\
$b= E \vee S \vee S^{\smile}$ \\
$u= P \vee O \vee M \vee D \vee S$ \\
$v= P \vee O \vee M \vee D^{\smile} \vee F^{\smile}$ \\
$n=E \vee F \vee D \vee O \vee S \vee F^{\smile} \vee D^{\smile} \vee
O ^{\smile} \vee S^{\smile}$ \\

\end{table*}

Figure \ref{PathConsistency} presents the pseudo code of the PC algorithm  \cite{van1996design} for IA networks. This algorithm\,(called also transitive closure algorithm) can easily be adapted to work with any QCN.  The procedure in Figure \ref{PathConsistency} starts by selecting, at each iteration, any three nodes $I$, $J$ and $K$ of the IA network and checking whether $R_{IK} = R_{IK} \cap (R_{IJ} \otimes R_{JK})$. If it is not the case, then $R_{IK}$ is updated and this update will be propagated to the rest of the network. The algorithm iterates until no more such changes are possible. $R_{IK}$\,(respectively $R_{IJ}$ and $R_{JK}$) is the binary constraint between events $I$ and   $K$\,(respectively between events $I$ and $J$; and between events $J$ and $K$). $\cap$ is the intersection operator between two relations\,(the result of the intersection between two relations is the common Allen primitives the two relations share). $\otimes$ is the composition operator between two relations, defined using the Allen composition table \cite{allen1983maintaining}. As we can notice, the main advantage of this PC algorithm is that only the triangle of edges whose labels have changed in the previous iteration need to be computed (as reported by Mackworth \cite{Mackworth77} and Allen \cite{allen1983maintaining}). This is achieved through the queue data structure ($L$) for maintaining the triangles that must be recomputed. The computation proceeds until the queue is empty. This will considerably reduce the number of triangles to be processed.

 The PC algorithm assumes that the constraint graph is complete. If the initial graph is not complete then it is transformed into a complete one by adding arcs labeled with the universal relation $I$.


\begin{algorithm} 
\caption{Path Consistency Algorithm for IA networks \cite{van1996design}.}
\label{PathConsistency}
\begin{algorithmic}[1]
\Procedure{PathConsistency} {}

	\State $PC \gets false$
	\State $L \gets \{(i,j) | 1 \leq i < j \le n \}$
	\While { $L \ne \phi$}
		\State choose and delete $(i,j)$ from $L$
		\For { $k \gets 1$ \textbf{to} $n$, $k \ne i$ and $k \ne j$ }
			\State $t \gets C_{ik} \cap C_{ij} \otimes C_{jk}$
			\If { $t \ne C_{ik}$ }
				\State $C_{ik} \gets t$
				\State $C_{ki} \gets Inverse(t)$				
				\State $L \gets L \cup \{(i,k)\}$
			\EndIf
			\State $t \gets C_{kj} \cup C_{ki} \otimes C_{ij}$
			\If { $t \ne C_{kj}$ }
				\State $C_{jk} \gets Inverse(t)$
				\State $L \gets L \cup \{(k,j)\}$
			\EndIf
		\EndFor
	\EndWhile

\EndProcedure
\end{algorithmic}
\end{algorithm}

In order to show how PC is  enforced on an IA network, let us consider the following tasks added to our Example \cite{Belouaer12}.
\noindent

\subsubsection*{Example\,2 }

\begin{quote}
{\em
\begin{enumerate}
\item Apply PC
\item Add the following information: John either {\bf started or arrived} just as Wendy {\bf started}
\item Apply PC

\item Add the following information: Mary and Wendy {\bf arrived together but started at different times}
\item Apply PC
\end{enumerate}
}
\end{quote}

\begin{figure}
    \centerline{\resizebox{95mm}{!}{\includegraphics{{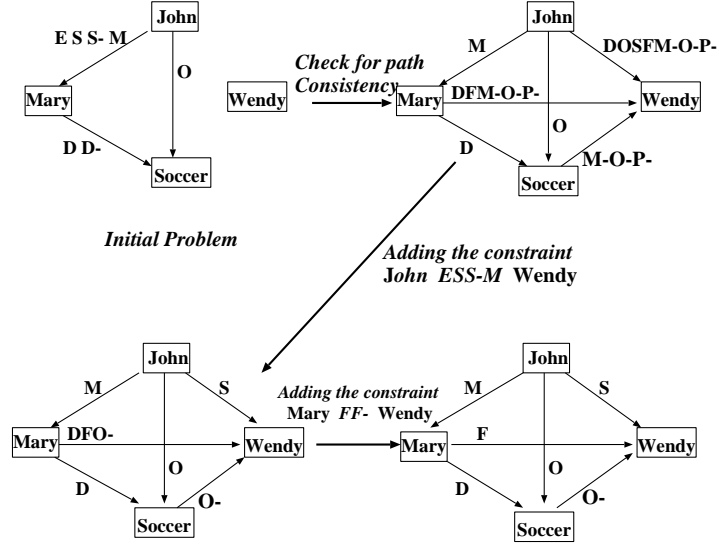}}}}
 \caption{Applying PC for the AI network in Example 2 \cite{DBLP:conf/appinf/Mouhoub03}}
  \label{DynPC}
\end{figure}

The above basically consists of applying PC and maintaining it every time a temporal information is added. Figure \ref{DynPC} illustrates the corresponding process on the corresponding IA network. After performing PC on the initial network (top left graph), we obtain a complete path consistent network. As we can notice, some of the primitives relations are removed (such as ``$E \vee S \vee S^{\smile}$'' between John and Mary) as they do not satisfy the path consistency. The second and third actions above are then applied and, as a result, we obtain the bottom left graph. Finally, applying the fourth and fifth tasks will result in the last graph (bottom right).\\

Partial Path Consistency (PPC) is an extension of PC based on chordal graphs. A chordal graph is a graph such that, for every four connected nodes and higher there must be one chordal edge \cite{bliek1999path}. The chordal graph is also called triangulated graph, given that it can be printed on a map where vertices are connected in triangle shapes. Partial Path Consistency assures that the chordal graph is consistent such as when applying regular PC \cite{bliek1999path}. Partial Path Consistency is much efficient than PC when working with chordal graphs. 

\section{Overview on Constraint Acquisition}
\label{sec:ConstraintAcquisition}

Constraint acquisition is the process of learning a target constraint network, from a set of examples. Several constraint acquisition algorithms have been proposed and the focus is to minimize the number of examples needed to converge to a target constraint network. In this regard, a passive learning algorithm, called CONACQ.1, has been proposed \cite{bessiere2007query}. CONACQ.1, is the passive version of the CONACQ acquisition system \cite{bessiere2015constraint}), and uses unit propagation to speed up the convergence process. In \cite{bessiere2013constraint}, a proposed active constraint acquisition technique, called QUACQ, works by asking the user to classify partial queries (those queries not involving all variables) in addition to membership queries (where the user is asked to answer ``yes'' or ``no'' depending on whether the provided assignment is a solution or not).  Partial queries are asked when given a negative example and allow QUACQ to converge in a number of queries that is logarithmic in the number of variables. An extension to this latter algorithm has been proposed in \cite{arcangioli2015} where multiple constraints are learned (and not only one as in QUACQ) using Minimal Unsatisfiability Subsets (MUS).

The pseudo code of QUACQ is shown in Algorithm \ref{QUACQAlgorithm}.

\begin{algorithm} 
\caption{QUACQ Algorithm \cite{bessiere2013constraint}}
\label{QUACQAlgorithm}
\begin{algorithmic}[1]
\Procedure{QUACQ} {}

\State $C_L \gets \phi$
\While { $true$ }
	\If {$sol(C_L) = \phi$}
		\State \Return ``collapse''
	\EndIf
	
	\State $e \gets$ example in $D^X$ accepted by $C_L$ and rejected by $B$

	\If {$e = nil$}
		\State \Return ``convergence on $C_L$''
	\EndIf
	\If {$Ask(e) = yes$}
		\State $B \gets B$ \textbackslash $ $ $_{KB}(e))$
	\Else
		\State $c \gets FindC(e,FindScope(e,\phi,X,false))$
		\If {$c = nil$}
			\State \Return ``collapse''
		\Else
			\State $C_L \gets C_L \cup \{c\}$
		\EndIf
	\EndIf
\EndWhile
\EndProcedure
\end{algorithmic}
\end{algorithm}


In \cite{bessiere2015constraint}, the active version of the CONACQ architecture, called CONACQ.2, is proposed  as shown in Algorithm \ref{CONACQAlgorithm}. CONACQ.2 starts with a bias $B$ (similar to our language set $\mathcal{B}$) and its corresponding background language $K$. $K$ is a set of declarative rules (definite Horn clauses) expressing some properties (such as the transitivity property) that can be used to enforce local consistency between constraints. In the case of QCNs, the set $K$ basically corresponds to the composition table we mentioned in the previous section. Given these two inputs $B$ and $K$, in addition to a given strategy to follow for generating the queries, the algorithm iterates by asking the user a new generated query at each time and expands the theory set $T$ (initially set to the empty set) according to the answer provided. More precisely, if the user answers ``no'' to a given query $q$ then the algorithm removes all the concepts that support this query (as shown in line 11 of the Algorithm, $K(q)$ corresponds here to the set of constrains $c$ violated by the query $q$). In case the answer is ``yes'' then all the concepts rejecting $q$ must be removed (as per line 13 of the Algorithm). The algorithm terminates when there is no query to generate. The target network, in the form of the clausal theory $T$, is then returned. Note that $a(c)$ denotes the Boolean atom corresponding to a given constraint $c$.

\begin{algorithm} 
\caption{CONACQ.2 \cite{bessiere2015constraint}}
\label{CONACQAlgorithm}
\begin{algorithmic}[1]
\Procedure{CONACQ.2} {}
\\
\textbf{Input:} a bias $B$, a background knowledge $K$, a strategy $Strategy$\\
\textbf{Output:} a clausal theory $T$ encoding the target network

\State $T \gets \emptyset$; $converged \gets false$; $N \gets \emptyset$
\While {$\neg converged$}
    \State $q \gets$  $QueryGeneration(B,T,K,N,Strategy)$
    \If {$q = nil$}
        \State $converged \gets true$
    \Else
        \If {$Ask(q) = no$}
            \State $T \gets T \wedge (\bigvee_{c \in K(q)} a(c))$
        \Else
            \State $T \gets T \wedge \bigwedge_{c \in K(q)} \neg a(c)$
        \EndIf
    \EndIf
\EndWhile

\Return $T$
\EndProcedure
\end{algorithmic}
\end{algorithm}

\section{Exact Learning of QCNs using Membership Queries}
\label{sec:ProposedAlgorithm}

Our proposed learning algorithm considers the following three cases, each with a different target QCN.

\begin{enumerate}
\item
	 Learning consistent and complete scenarios: the target QCN is a complete graph where each edge (constraint) contains exactly 1 relation (Allen primitive in the case of IA networks).
\item

	 Learning consistent but incomplete scenarios: same as case 1 but in the case of incomplete graphs (some constraints do not exist between pairs of variables).

\item Learning QCNs: the target QCN is a graph (can be complete) where each edge contains 1 or more primitives (each constraint has one or more relations). Note that, in this particular case we are learning a QCN problem rather than a consistent scenario (solution) as it is in cases 1 and 2.
\end{enumerate}

Following CONACQ.2 \cite{bessiere2015constraint}, our algorithm learns through membership queries. PC is used to reduce the total number of queries needed to learn the target QCN. Note that when using PC for case 3, our learning algorithm will return a path consistent QCN problem (instead of a consistent scenario as in cases 1 and 2). PC is enforced using the algorithm in Figure \ref{PathConsistency}, with the following improvement.

\begin{itemize}
\item We changed as reported in \cite{Bessiere96} the way composition and intersection of relations are achieved during the path consistency process\,(following the principle ``one support is sufficient'').
\end{itemize}




\subsection{Proposed Learning Algorithm}
\label{case1}

Our algorithm starts with all the constraints completely unknown (corresponding to the universal relation $I$ in the case of IA networks). The user is then asked a membership query for each relation within each constraint.   If the answer is ``yes'' for a given query, the corresponding constraint will be replaced by the relation that has been confirmed (all the other relations will be eliminated). Otherwise (if the answer is ``no''), the related relation will be removed from the constraint.   After every query, PC is applied on the graph to remove path inconsistent relations due to this recent update.  This will reduce the number of relations per constraint which will  reduce the number of subsequent queries and helps getting the target QCN scenario (solution) sooner.

The pseudo-code of our method is listed in Algorithm \ref{LocalConsistencyAlgorithm}. $G_t$  is initially set to a complete constraint graph with universal relations. $\mathcal{B}$ and $CT$ are respectively the set of possible relations and composition table of the QCN to learn. $QueryGeneration$ is a function that generates a membership query each time it is called. We use two implementations of this function. In the first one, the constraint and its related relation (to confirm) are picked randomly. In the second implementation, the relations to confirm are picked according to the fail-first principle used when solving general Constraint Satisfaction Problems (CSPs) \cite{Dechter:2003:CP:861888}. In this regard, we use the  ordering heuristics proposed in \cite{van1996design}, namely ``weight'' and ``cardinality'' in the case of IA networks. The idea behind the ``weight'' heuristic is to quantify the restriction imposed by a relation, when assigned to a given edge, on the temporal constraints of the other edges. The ``cardinality'' heuristic is a special case of the ``weight'' heuristic when considering that all relations have the same weight (equal to 1).
The algorithm iterates by processing a query at each time, until a solution (target QCN) is found or a path inconsistency is detected.

\begin{algorithm} 
\caption{Learning a QCN for Case 1}
\label{LocalConsistencyAlgorithm}
{\relsize{0}
\begin{algorithmic}[1]
\Procedure{LearningQCN} {}\\
\textbf{Input:} : a language set $\mathcal{B}$, a composition table $CT$\\
\textbf{Output:} a target QCN $G_t$
\State $ G_t \gets$ complete graph with universal relations
\State $q \gets QueryGeneration(G_t)$
\While { $q \neq nil$ }	
        \State $r \gets Relation(q)$
		\If {$Ask(q) = ``yes''$}
			\State $ConfirmRelation(G_t,r)$
		\Else
			\State $RemoveRelation(G_t,r)$
		\EndIf
		\State $status \gets PC(G_t,CT)$
		\If {$status = ``inconsistent''$}
			\State \Return $``collapse''$
		\EndIf
        \State $q \gets QueryGeneration(G_t)$
\EndWhile
\State \Return $G_t$
\EndProcedure
\end{algorithmic}
}
\end{algorithm}

\label{case2}


 Given that the constraint graph is not necessarily complete, our algorithm for case 2 differs from the previous one, as follows. If the answer is ``yes'' for a given query, the relation is confirmed but we do not remove the other relations as the corresponding constraint can be the universal relation, $I$. Instead, we ask the user a second query for the same constraint in order to check if this latter is $I$. If the answer is ``yes'' then the constraint is confirmed to be universal, otherwise (the answer is``no'') we replace the constraint by the relation confirmed with the first query. The rationale is that, in case 2 we either have a single relation or a universal relation for each constraint.

\label{case3}

In case 3, given that the number of relations per constraint varies from 1 to $|\mathcal{B}|$ (13 in the case of IA networks) then we will have to ask the user subsequent queries for the same constraint regardless of the answer. More precisely, if the answer is ``yes'' then we   need to ask the user for the remaining relations as we can have more than one relation per constraint.

\subsection{Dealing with  Inconsistent Answers}
\label{mistake}

As stated  in Algorithm \ref{LocalConsistencyAlgorithm}, our learning method collapses (fails to return the target QCN) if PC detects an inconsistency after one of the constraints becomes empty (has no relations). In this case, we need to identify the query that has been answered incorrectly. We address this task using a backtrack search algorithm to go backward and let the user confirm previous answers until we reach the state where the user changes the answer of a query. Given that we can have more than one incorrect answer, every time a response to a query is changed, we resume the normal querying starting back from the state where the user fixed the incorrect answer. For example, assume that after the query $q_j$ is answered, an inconsistency is detected. Our algorithm will then backtrack until it identifies the query $q_m$ causing this inconsistency.   The user will then change the answer and our learning algorithm will resume  from query $q_{m+1}$. The pseudo code of our method is listed in Algorithm \ref{BacktrackingLocalConsistencyAlgorithm}. This algorithm is very similar to Algorithm \ref{LocalConsistencyAlgorithm} except that we save the query and the current $G_t$ at each time in stack $s$. 

\begin{algorithm} 
\caption{Learning a QCN with Mistakes}
\label{BacktrackingLocalConsistencyAlgorithm}
{\relsize{-1}
\begin{algorithmic}[1]
\Procedure{LearningWithMistakesQCN} {}\\
\textbf{Input:}  $\mathcal{B}$: Language set. $CT$: Composition Table\\
\textbf{Output:} $G_t:$ target QCN
\State $ G_t \gets$ complete graph with universal relations
\State $s \gets \emptyset$
\State $q \gets QueryGeneration(G_t)$
\State $ stackpush(s,<G_t,q>)$
\While {$q \neq nil$ }	
        \State $r \gets Relation(q)$
		\If {$Ask(q) = ``yes''$}
			\State $ConfirmRelation(G_t,r)$
		\Else
			\State $RemoveRelation(G_t,r)$
		\EndIf
		\State $status \gets PC(G_t,CT)$
		\While {$status = ``inconsistent''$}
            \State $ stackpop(s,<G_t,q>)$
            \If {$Ask(q) = ``yes''$}
			\State $ConfirmRelation(G_t,r)$
		    \Else
			\State $RemoveRelation(G_t,r)$
		    \EndIf
		\State $status \gets PC(G_t,CT)$
		\EndWhile
        \State $q \gets QueryGeneration(G_t)$
        \State $ stackpush(s,<G_t,q>)$
\EndWhile
\State \Return $G_t$
\EndProcedure
\end{algorithmic}
}
\end{algorithm}

\section{Analyzing the Complexity of Learning QCNs}
\label{sec:TheoreticalLimits}

One of the major issues when learning a structure is analyzing the information complexity of the best possible learning algorithm. In our settings, this corresponds to the smallest possible number of queries to exactly learn an arbitrary path consistent QCN.

We adopt a learning paradigm known as exact identification from membership queries \cite{Angluin:1988:QCL:639961.639995} where we fix an input space $X$ and define a set of concepts $\mathcal{F}:X\rightarrow \{0,1\}^{|X|}$. The learning algorithm (i.e., learner)  receives information in the form of $(x_1,\ell_1)$, $(x_2,\ell_2)\dots, (x_m,\ell_m)$ where $x_i\in X$ and $\ell_i\in \{0,1\}$ and asked to exactly identify the target concept $f^*\in \mathcal{F}$ where $f^*(x_i)=\ell_i$ for all $i\in \{1,\dots,m\}$. Thus, the goal is to identify $f^*$ exactly out of all other concepts in the class.

For QCNs, we are interested in analyzing the learnability of the set of complete QCNs, the set of incomplete QCNs, and the set of both complete and incomplete QCNs. In particular, we are provided with a set  of $n$ entities $I$ and $p$ relations $R$ and an example is defined to be a pair of entities with a relation i.e., $(a,b,r)$ such that $i,j\in I$ and $r\in R$. The input space $X$ is defined in an obvious way $X=\{(a,b,r)\ |\ a\neq b\in I$ and $r\in R\}$ with size of $n(n-1)p/2$.  For a set of QCNs $S$, a concept $f$ is representable by $S$ if there exists a QCN $Q\in S$ such that $f(x)=1$ if and only if $x$ holds in $Q$. The concept class $\mathcal{F}$ is then defined as the set of all concepts that are representable by $S$.

We consider the three concept classes $\mathcal{F}^c,\bar{\mathcal{F}}$ and $\mathcal{F}^*$ that represent respectively the set of concepts that are representable by the set of complete, incomplete, and both complete and incomplete QCNs.

The complexity of \emph{teaching} a concept $f$ can be characterized as the minimum number of instances that distinguish $f$ from all other concepts in the class $\mathcal{F}$ \cite{GK95}. This is also known as the teaching dimension of $f$ i.e., $\TDim(f,\mathcal{F})$ A class $\mathcal{F}$ has a teaching dimension $d$ iff there exists a concept $f\in \mathcal{F}$ with $\TDim(f,\mathcal{F})=d$ but there exists no $f'\in \mathcal{F}$ such that $\TDim(f',\mathcal{F})>d$. In other wods, teaching dimension of a class equals the teaching dimension of the hardest concept in the class.

We analyze the teachability of $QCNs$ and find $\TDim$ for the three concept classes of QCNs. It is an established fact in computational learning theory that the complexity of teaching, i.e., $\TDim$ of a class lower bounds the number of queries required to exactly learn a concept. Thus, if we happen to know the teaching dimension of a class $\mathcal{F}$ is $d$ then we are certain there exists no algorithm $\mathcal{A}$ that would require less than $d$ queries. Therefore, our next results shed some lights into the number of queries required to learn QCNs.

\begin{proposition}
	$\TDim(\mathcal{F}^c)=n(n-1)/2$.
\end{proposition}
\begin{proof}
	Every concept $f$ in $\mathcal{F}^c$ has $n(n-1)/2$ edges as it is a complete QCN, and every edge $(a,b,r)$ can be taught individually by one example $(a,b,r)$. This establishes the fact that $\TDim(\mathcal{F}^c)\geq n(n-1)/2$ as there exists $n(n-1)/2$ many edges in any complete QCN.
	
It remains to show a concept $f\in \mathcal{F}^c$ that cannot be taugh with less than $n(n-1)/2$ examples. Consider the concept $f$ where we cannot extract any new relation between $(a,c)$ for any two existing edges $(a,b)$ and $(b,c)$. To teach $f$, we require to provide the relation of every edge in $f$ which needs $n(n-1)/2$ examples.

\end{proof}

For the class of incomplete QCNs, it is harder to teach compared to the set of complete QCNs, as there are more complex networks that are repsentable in this class. This intuition is indeed valid as the following result states.
\begin{proposition}
	$\TDim(\bar{\mathcal{F}})=n(n-1)$.
\end{proposition}
\begin{proof}
For an arbitrary pair $(a,b)$, a two instances $x=(a,b,r)$ and $x'=(a,b,r')$ where $r\neq r'$ can be used to teach the relation. Specifically, when teaching any concept $f$, $f(x)=f(x')=1$ if there exists no edge between $a$ and $b$ otherwise $f(x)=1$ and $f(x')=0$ where $r$ is the relation between $i$ and $j$. This settles the teaching dimension of an artbitrary concept to be less than or equal $n(n-1)$.

To show that $\TDim(\bar{\mathcal{F}})=n(n-1)$, consider the concept with no edges (i.e., all relations are universal relations). This needs two instances per pair $(a,b)$ and thus we would need  $n(n-1)$ instances to teach it.

\end{proof}

\begin{proposition}
	$\TDim(\mathcal{F}^*)=|X|$
\end{proposition}
\begin{proof}
	Consider again the cocept $f$ with no edges. We cannot teach $f$ in less than $|X|$ examples. Assume there exists a teaching set $T$ of size $|X|-1$ to teach $f$. Let $X=T\cup {(a,b,r)}$. There exists another concept $f'\in \mathcal{F}^*$ with relations identical to $f$ except that for the pair $(a,b)$ it has the relation $R\backslash r$. $T$ is also consistent with $f'$ which contradicts with $T$ being a teaching set for $f$. This, along with the fact that $\TDim(\mathcal{F}^*)$ is at most $|X|$, prove our claim.
	
\end{proof}

\section{Experimentation}
\label{sec:Experimentation}


We report on the experiments conducted in order to assess the effect of PC on the number of queries needed to learn a QCN network. All the experiments are conducted on a Dell XPS 8900, i7-6700K, 32 GB RAM, running Linux. All the algorithms are coded in Java.

We compare variants of our proposed learning algorithm to CONACQ.2, for each of the three cases listed in Section \ref{sec:ProposedAlgorithm}. In the case of scenarios with incorrect answers (as discussed in Section \ref{mistake} above), variants of our algorithm are compared to each other in order to assess the effect of path consistency and queries ordering heuristics on time needed to deal with incorrect answers, in order to successfully converge to the target QCN. The variants of our algorithm include the case where path consistency is used, with or without a given ordering heuristic. If path consistency is used, we consider two situations: the case where the $QueryGeneration$ function generates the queries randomly as well as using an ordering heuristic as described in Section \ref{sec:ProposedAlgorithm}. Moreover, for case two we consider PPC in addition to PC, given that we have incomplete graphs.

\subsection{Experiments on IA constraint network instances}
\label{IAinstance}

The first set of experiments are conducted on random IA constraint network instances,  generated as follows. We first produce a random consistent temporal scenario (that we call $G_t$). We then add primitives randomly to get an initial graph $G_{problem}$. The goal here is to achieve the following, at the end of the learning process: $G_{problem} = G_t$. Each query is generated with a probability $P_y$ of getting a ``yes'' answer.  The general approach is similar for all the cases we consider in the following, and differs in terms of the target $G_t$. 

In case 1, we use the $\mathbf{S}(n, p)$ model \cite{van1996design} to generate the random instances. More precisely, we start by generating $n$ numeric random intervals (pairs of natural values) assigned to each temporal variable in the graph. Allen primitives are then deduced from pairs of these numeric intervals. For example, let us assume the following assignments to two temporal events $X$ and $Y$: $X=(3, 11)$ and $Y=(7, 18)$. The corresponding Allen primitive is then: $(X\; O\; Y)$ i.e. $X$ overlaps $Y$. The model $\mathbf{S}(n, p)$ ensures a consistent solution since all variables have numeric intervals. A random consistent scenario is then build and is considered as the target graph $G_t$. Our algorithm then starts from a complete graph $G_{problem}$ where each edge contains the universal relation $I$ (the disjunction of all the 13 primitives). This basically means that we start from a completely unknown QCN to learn.\\

Case 2 is similar to case 1 with the following difference when generating the target network. Some edges in $G_t$ will be removed based on a parameter $P_u$ corresponding to the percentage of universal relations in $G_t$. These relations (edges to be removed) will be in stored in a set $E_u$ that is used in the query generation.  $Q_{solution}$ is generated by traversing all the edges in $G_t$, and by adding all the universal relations in $E_u$.

 For case 3, we first generate $G_t$ as done in case 1. We then randomly add more primitives to each constraint $(X,Y)$, in order to form a randomly generated consistent QCN corresponding to a complete graph ($G_t$).  $Q_{solution}$ contains queries for the primitives within the constraint $(X,Y)$.

\begin{figure}[ht]
{
\begin{center}

\begin{tikzpicture}

\begin{axis}[
  legend style={font=\tiny},
     legend style={at={(0.98,0.9)},anchor=north east},
	scale=1.25,
 xlabel=Case 1: Percentage of ``yes'' answers,
 ylabel=Number of Queries]
\addplot table [y=$Naive$, x=$percent$]{case1bis2.dat};
\addlegendentry{$CONACQ.2$}
\addplot table [y=$PC$, x=$percent$]{case1bis.dat};
\addlegendentry{$PC$}
\addplot table [y=$PCS1$, x=$percent$]{case1bis.dat};
\addlegendentry{PC with ``cardinality'' ordering heuristic }
\addplot table [y=$PCS2$, x=$percent$]{case1bis.dat};
\addlegendentry{PC with ``weight'' ordering heuristic}
\addplot table [y=$PCS3$, x=$percent$]{case1bis.dat};
\addlegendentry{PC with ``cardinality descending'' ordering heuristic}
\end{axis}
\end{tikzpicture}
	\caption{Test results on random IA instances for case 1}
	\label{Figcase1}
\end{center}
}
\end{figure}
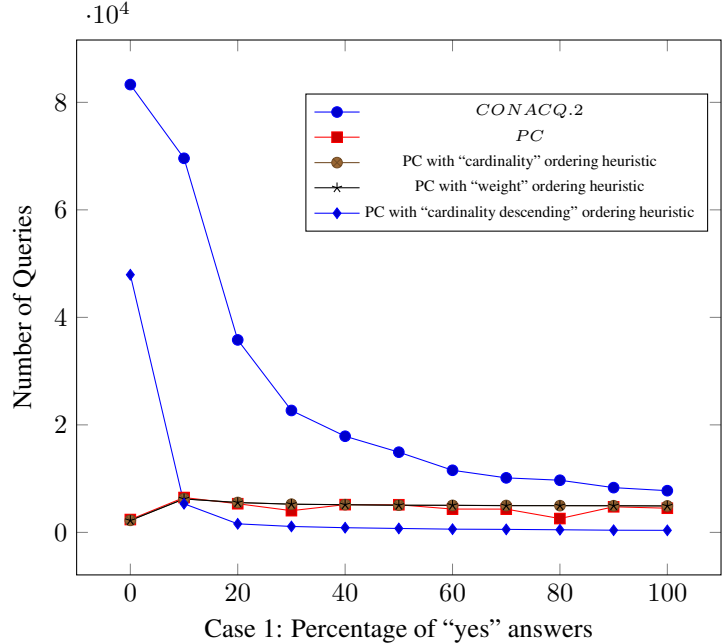

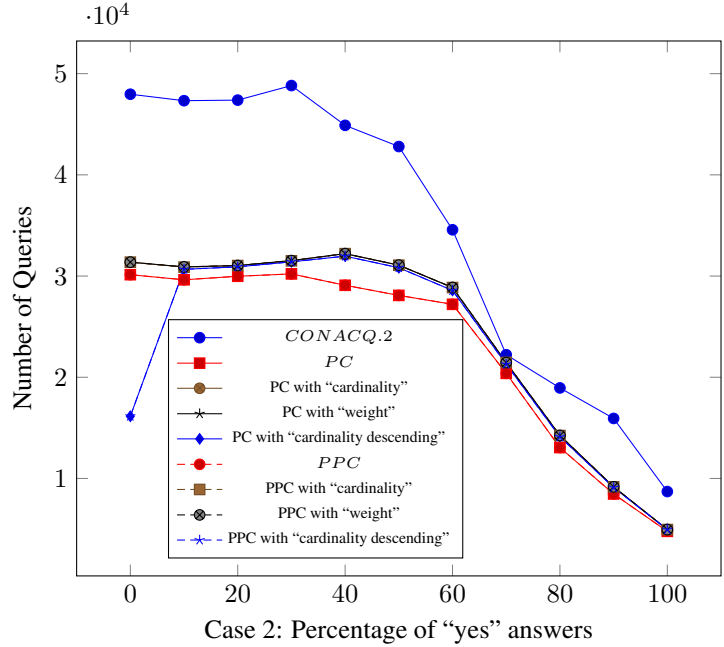
\begin{figure}[ht]
\begin{center}
\begin{tikzpicture}
\begin{axis}[
  legend style={font=\tiny},
     legend style={at={(0.6,0.48 )},anchor=north east},
	scale=1.25,
  xlabel=Case 2: Percentage of ``yes'' answers,
  ylabel=Number of Queries]
\addplot table [y=$Naive$, x=$percent$]{case2bis2.dat};
\addlegendentry{$CONACQ.2$}
\addplot table [y=$PC$, x=$percent$]{case2bis.dat};
\addlegendentry{$PC$}
\addplot table [y=$PCS1$, x=$percent$]{case2bis.dat};
\addlegendentry{PC with ``cardinality''    }
\addplot table [y=$PCS2$, x=$percent$]{case2bis.dat};
\addlegendentry{PC with ``weight''  }
\addplot table [y=$PCS3$, x=$percent$]{case2bis.dat};
\addlegendentry{PC with ``cardinality descending''   }
\addplot table [y=$PPC$, x=$percent$]{case2bis.dat};
\addlegendentry{$PPC$}
\addplot table [y=$PPCS1$, x=$percent$]{case2bis.dat};
\addlegendentry{PPC with ``cardinality''     }
\addplot table [y=$PPCS2$, x=$percent$]{case2bis.dat};
\addlegendentry{PPC with ``weight''   }
\addplot table [y=$PPCS3$, x=$percent$]{case2bis.dat};
\addlegendentry{PPC with ``cardinality descending''  }
\end{axis}
\end{tikzpicture}
	\caption{Test results on random IA instances for case 2}
	\label{Figcase2}
\end{center}
\end{figure}

\begin{figure}[ht]
\begin{center}
\begin{tikzpicture}
\begin{axis}[
  legend style={font=\tiny},
     legend style={at={(0.95,0.85 )},anchor=north east},
	scale=1.25,
  xlabel=Case 3: Percentage of ``yes'' answers,
 ylabel=Number of Queries]
\addplot table [y=$Naive$, x=$percent$]{case3bis2.dat};
\addlegendentry{$CONACQ.2$}
\addplot table [y=$PC$, x=$percent$]{case3bis.dat};
\addlegendentry{$PC$}
\addplot table [y=$PCS1$, x=$percent$]{case3bis.dat};
\addlegendentry{PC with ``cardinality''    }
\addplot table [y=$PCS2$, x=$percent$]{case3bis.dat};
\addlegendentry{  with ``weight''   }
\addplot table [y=$PCS3$, x=$percent$]{case3bis.dat};
\addlegendentry{  with ``cardinality descending''   }
\end{axis}
\end{tikzpicture}
	\caption{Test results on random IA instances for case 3}
	\label{Figcase3}
\end{center}
\end{figure}
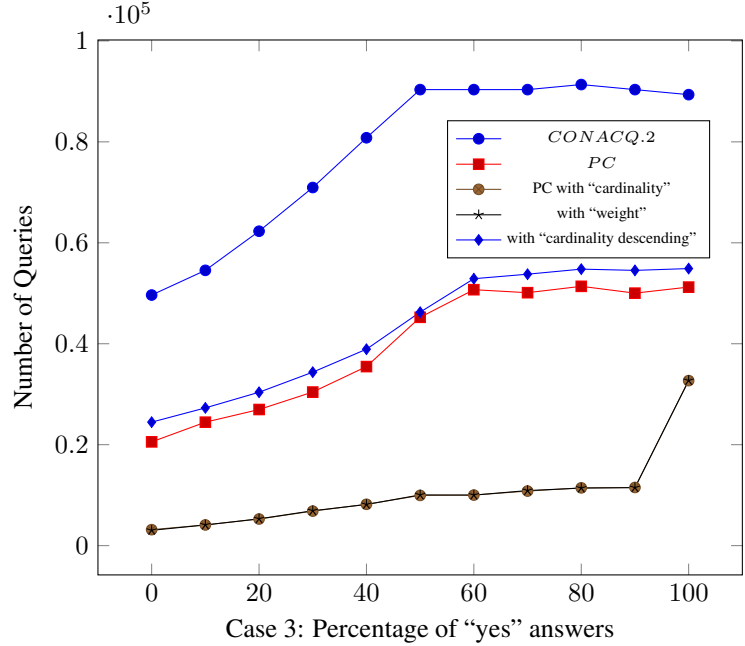

 Figures \ref{Figcase1}, \ref{Figcase2}, and \ref{Figcase3} show the comparative results for the three cases when 100 variables are considered. It is clear from the chart corresponding to case 1 that PC has a significant effect on reducing the number of queries especially when there is a small percentage of ``yes'' answers. For instance, in the extreme case where there is no ``yes'' answers, CONACQ.2 requires more than 80000 queries while the learning method using PC only needs about 2500 queries to reach the target IA scenario. By increasing the percentage of ``yes'' answers, the number of queries starts to drop down to 7000 queries for $100\%$ of ``yes'' answers using CONACQ.2, which is still far from the 3731 queries that are asked using PC. The ``cardinality descending'' heuristic is the most effective one as we can easily see from the chart, while the other two do not seem to have an effect when compared to PC without heuristics.

The situation in case 2 is similar to the one in case 1. All the methods using path consistency are better than CONACQ.2. These methods have however similar performance.

 For case 3, we can easily see that there is a significant difference between the 3 methods considered. PC with ``cardinality'' ordering heuristic is the winner in this situation and is followed by the PC method without heuristic. These results are justified by the fact that we are dealing with complete graphs with much more relations per constraint than cases 1 and 2.


In order to simulate the situations involving incorrect answers, the query generation for inconsistent scenarios is produced as follows. Mistakes are picked randomly with a percentage ratio $P_m$.   Query generators are similar to the query generators in case 1, case 2, and case 3. However, queries are labelled with $isMistake$ as a boolean indicator which is checked when answering each query. If $isMistake$ is true, then the answer would be incorrect, otherwise, the answer must be correct.

Tables \ref{Inconsistenttbl1}, \ref{Inconsistenttbl2} and \ref{Inconsistenttbl3} show the results for inconsistent scenarios due to mistakes for each of the 3 cases when 100 variables are considered. Here. BT refers the baseline backtrack search algorithm (with no path consistency and no queries ordering heuristics).  As we can notice, in all cases the number of queries, mistakes and related backtracks is considerably reduced when path consistency is used.

In Table \ref{Inconsistenttbl1}, we can easily see that there is a significant difference between BT and PC in terms of number of queries, number of mistakes detected and running time. There are 14 times more mistakes detected by BT than the PC method.  This is explained by the fact that PC removes inconsistent relations at each time leaving less chances for the user to choosing them. Also, the ordering heuristic does not seem to do a good job in this case.

In Table \ref{Inconsistenttbl2}, we notice that the performance PPC is better than the one of PC and this is due to the fact that we are dealing with incomplete graphs. Also, the ``weight'' ordering heuristic is effective in this case for both PC and PPC. This is again explained by the nature of problems we are dealing with in case 2.

Table \ref{Inconsistenttbl3} clearly shows that both the ``cardinality'' and ``weight'' ordering heuristics are effective given that the objective is to produce a path consistent QCN.

\begin{table}[H]
{\relsize{-2}
	\begin{center}
\caption{Results for learning IA scenarios with mistakes: case 1}
\label{Inconsistenttbl1}
	\begin{tabular}{ | c | c | c | c | c | }
		\hline
		Method & \# of Queries & Time(s) & Mistakes & Backtracks \\
		\hline
BT	&		1623776 &	968	&	399 &	398 \\
\hline
PC	&		4444 &		54	&	28 &	28 \\
\hline
PC with ``cardinality''	&		6879 &		87	&	43 &	43 \\
\hline
PC with ``weight'' 	&		7819 &		126	&	36 &	36 \\
\hline
PC with ``cardinality descending'' 	&		38367 &		532	&	75 &	75 \\
\hline
	\end{tabular}
	\end{center}
	
}	
\end{table}

\begin{table}[H]
{\relsize{-2}
	\begin{center}
\caption{Results for learning IA scenarios with mistakes: case 2}
\label{Inconsistenttbl2}
	\begin{tabular}{ | c | c | c | c | c | }
		\hline
		Method & \# of Queries & Time(s) & Mistakes & Backtracks \\
		\hline
BT &		63206804&	3430.845 &	6676 &	6666 \\
\hline
PC &			36662	&	458.491 &		4 &		5 \\
\hline
PPC &			10598	&	114.350 &		3 &		2 \\
\hline
PC with ``cardinality'' &		28906	&	327.168 &		5 &		0 \\
\hline
PPC with ``cardinality'' &		88593	&	986.297 &		14 &		14 \\
\hline
PC with ``weight''  &		8786	&	198.325 &		1 &		1 \\
\hline
PPC with ``weight''  &		8655	&	109.046 &		2 &		2 \\
\hline
PC with ``cardinality descending''  &		132225	&	2470.913 &	19 &		20 \\
\hline
PPC with ``cardinality descending''  &		67066	&	706.155 &		8 &		9 \\
\hline
	\end{tabular}
	\end{center}
	
}	
\end{table}

\begin{table}[H]
{\relsize{-2}
	\begin{center}
	\caption{Results for learning IA networks with mistakes: case 3}
	\label{Inconsistenttbl3}
	\begin{tabular}{ | c | c | c | c | c | }
		\hline
		Method & \# of Queries & Time(s) & Mistakes & Backtracks \\
		\hline
BT &			2384321 &		123.455 &		66 &		66 \\
\hline
PC &			40917 &		542.344 &		41 &		1 \\
\hline
PC with ``cardinality'' &			11514 &		46.772 &		10 &		6 \\
\hline
PC with ``weight''  &			11280 &		54.366 &		11 &		2 \\
\hline
PC with ``cardinality descending''  &			330215 &		5323.526 &	69 &		61 \\
\hline
	\end{tabular}
	\end{center}

}
\end{table}

\subsection{Experimentation on RCC 8 instances}

The second set of experiments are conducted on Region Connection Calculus (RCC) \cite{LI2003121} constraint network instances, randomly generated as follows. Starting from a set of start and end points randomly generated on a 1 D map, we extract the corresponding RCC 8 relations, which will form the random consistent scenario ($G_t$). We then add RCC 8 primitives randomly to get an initial graph $G_{problem}$. The rest of the process is similar to the one described in Section \ref{IAinstance}.

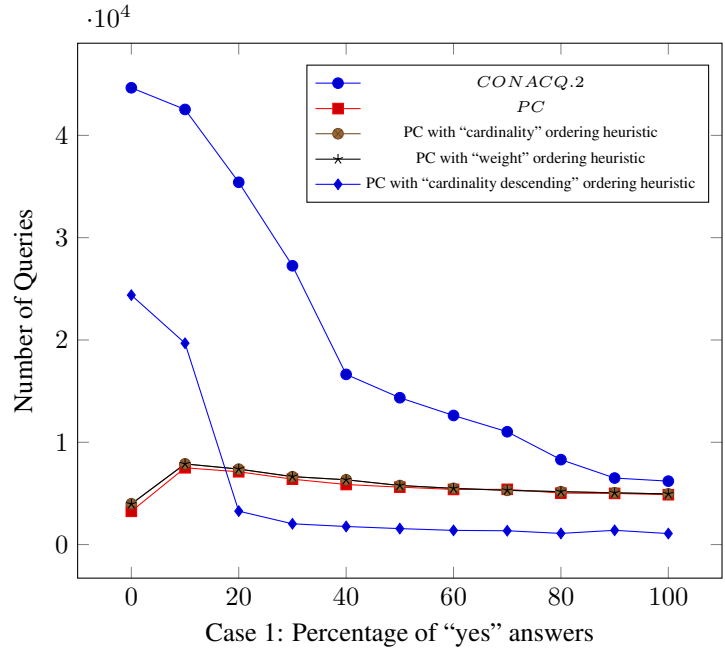
\begin{figure}
\begin{center}
\begin{tikzpicture}
{
\begin{axis}[
  legend style={font=\tiny},
     legend style={at={(0.98,0.96)},anchor=north east},
	scale=1.25,
 xlabel=Case 1: Percentage of ``yes'' answers,
 ylabel=Number of Queries]
\addplot table [y=$Naive$, x=$percent$]{case1STNbis2.dat};
\addlegendentry{$CONACQ.2$}
\addplot table [y=$PC$, x=$percent$]{case1STNbis2.dat};
\addlegendentry{$PC$}
\addplot table [y=$PCS1$, x=$percent$]{case1STNbis.dat};
\addlegendentry{PC with ``cardinality'' ordering heuristic }
\addplot table [y=$PCS2$, x=$percent$]{case1STNbis.dat};
\addlegendentry{PC with ``weight'' ordering heuristic}
\addplot table [y=$PCS3$, x=$percent$]{case1STNbis.dat};
\addlegendentry{PC with ``cardinality descending'' ordering heuristic}
\end{axis}
}
\end{tikzpicture}
	\caption{Test results for random STNs: case 1}
	\label{Figcase1STN}
\end{center}
\end{figure}

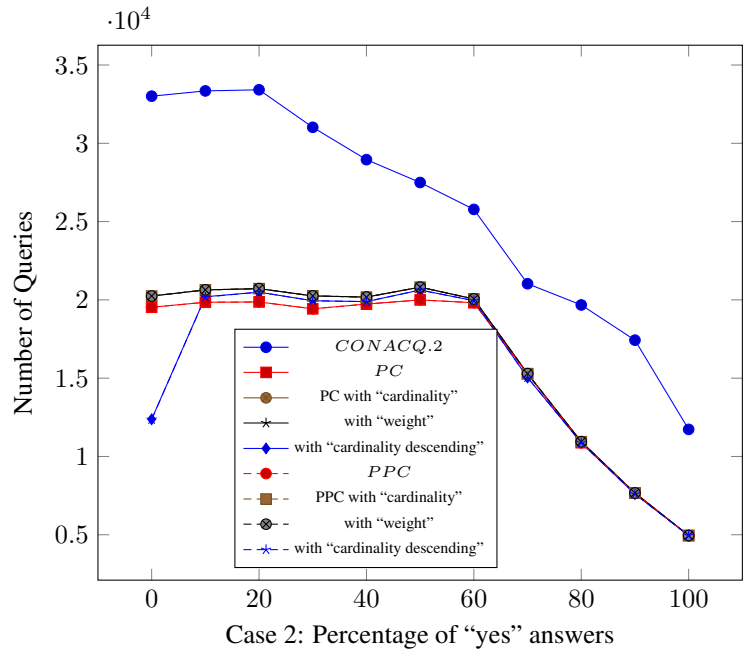
\begin{figure}
\begin{center}
\begin{tikzpicture}
\begin{axis}[
  legend style={font=\tiny},
     legend style={at={(0.62,0.47)},anchor=north east},
	scale=1.25,
  xlabel=Case 2: Percentage of ``yes'' answers,
  ylabel=Number of Queries]
\addplot table [y=$Naive$, x=$percent$]{case2STNbis2.dat};
\addlegendentry{$CONACQ.2$}
\addplot table [y=$PC$, x=$percent$]{case2STNbis.dat};
\addlegendentry{$PC$}
\addplot table [y=$PCS1$, x=$percent$]{case2STNbis.dat};
\addlegendentry{PC with ``cardinality''    }
\addplot table [y=$PCS2$, x=$percent$]{case2STNbis.dat};
\addlegendentry{  with ``weight''  }
\addplot table [y=$PCS3$, x=$percent$]{case2STNbis.dat};
\addlegendentry{  with ``cardinality descending''   }
\addplot table [y=$PPC$, x=$percent$]{case2STNbis.dat};
\addlegendentry{$PPC$}
\addplot table [y=$PPCS1$, x=$percent$]{case2STNbis.dat};
\addlegendentry{PPC with ``cardinality''     }
\addplot table [y=$PPCS2$, x=$percent$]{case2STNbis.dat};
\addlegendentry{ with ``weight''   }
\addplot table [y=$PPCS3$, x=$percent$]{case2STNbis.dat};
\addlegendentry{ with ``cardinality descending''  }
\end{axis}
\end{tikzpicture}
	\caption{Test results for random STNs: case 2}
	\label{Figcase2STN}
\end{center}
\end{figure}

\begin{figure}
\begin{center}
\begin{tikzpicture}
\begin{axis}[
  legend style={font=\tiny},
     legend style={at={(0.96,0.28)},anchor=north east},
	scale=1.25,
  xlabel=Case 3: Percentage of ``yes'' answers,
 ylabel=Number of Queries]
\addplot table [y=$Naive$, x=$percent$]{case3STNbis2.dat};
\addlegendentry{$CONACQ.2$}
\addplot table [y=$PC$, x=$percent$]{case3STNbis2.dat};
\addlegendentry{$PC$}
\addplot table [y=$PCS1$, x=$percent$]{case3STNbis2.dat};
\addlegendentry{PC with ``cardinality''    }
\addplot table [y=$PCS2$, x=$percent$]{case3STNbis2.dat};
\addlegendentry{ with ``weight''   }
\addplot table [y=$PCS3$, x=$percent$]{case3STNbis2.dat};
\addlegendentry{  with ``cardinality descending''   }
\end{axis}
\end{tikzpicture}
	\caption{Test results for random STNs: case 3}
	\label{Figcase3STN}
\end{center}
\end{figure}
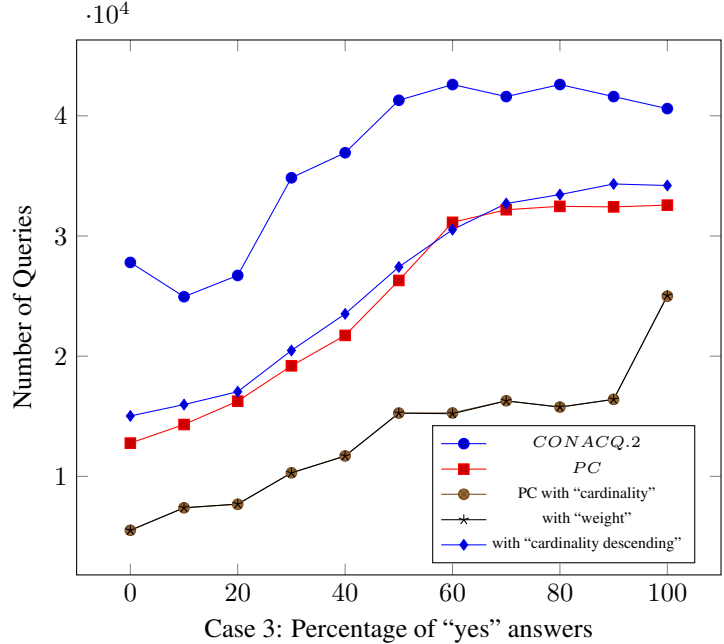

Figures \ref{Figcase1STN}, \ref{Figcase2STN}, and \ref{Figcase3STN} show the comparative results for the 3 cases when 100 variables are considered. As we can easily see from the three charts, PC has a considerable effect on reducing the number of queries. This is particularly noticeable in Figure \ref{Figcase1STN} where there is a small percentage of ``yes'' answers. Like for IA networks, the ``cardinality descending'' heuristic is the most effective one for case 1. While there is no obvious winner for case 2, PC with ``cardinality''is the best for case 3, and this is explained by the fact that we are dealing here with complete graphs.

Tables \ref{case1IncSTN}, \ref{case2IncSTN} and \ref{case3IncSTN} show the results for inconsistent scenarios due to mistakes for each of the 3 cases when 100 variables are considered. As for IA networks, in all cases the numbers of queries, mistakes and related backtracks are considerably reduced when path consistency is used. In Table \ref{case1IncSTN}, there are more than 10 times mistakes detected by BT than the PC method.  This is explained by the fact that PC detects and removes many inconsistent relations at each time leaving less chances for the user to choosing them. We also notice from the table that the ``weight'' ordering heuristic is the best one to do the job. Otherwise, the plain PC algorithm is better than the one with the other two heuristics. In Table  \ref{case2IncSTN}, we notice that the performance  PC is slightly better than PPC. Also, the ``weight'' ordering heuristic is effective in this case for both PC and PPC. This is again explained by the nature of problems we are dealing with in case 2. Table \ref{case3IncSTN} clearly shows that both   ``cardinality'' and ``weight'' (followed by ``cardinality descending'') ordering heuristics are effective given that the objective is to produce a path consistent STNs.



\begin{table}[htbp]
{\relsize{-2}
 
  \caption{Results for learning STNs with mistakes: case 1}
 \begin{center}
    \begin{tabular}{ | c | c | c | c | c | }
    \hline
          &  {\# of Queries} &  {Time(s)} &  {Mistakes} & {Backtracks} \\
     \hline
    BT & 4481736 & 167.378 & 260   & 247 \\
    PC    & 9047  & 546.513 & 23    & 23 \\
    PC with ``cardinality''  & 13203 & 362.221 & 43    & 43 \\
    PC with ``weight''  & 6935  & 220.029 & 32    & 32 \\
    PC with ``cardinality descending''  & 60793 & 4132.875 & 85    & 85 \\
    \hline
    \end{tabular}%
    \end{center}
	\label{case1IncSTN}
}
\end{table}%

\begin{table}[htbp]
{\relsize{-2}
 \begin{center}
  \caption{Results for learning STNs with mistakes: case 2}

    \begin{tabular}{ | c | c | c | c | c | }
      \hline
          &  {\# of Queries} &  {Time(s)} & \multicolumn{1}{l}{Mistakes} & {Backtracks} \\
          \hline
    BT & 14329123 & 443.424 & 2853  & 2846 \\
    PC    & 43811 & 1754.507 & 6     & 7 \\
    PPC   & 54229 & 1388.253 & 7     & 8 \\
    PC with ``cardinality''  & 130068 & 6257.959 & 16    & 15 \\
    PPC with ``cardinality'' & 133498 & 4014.92 & 16    & 17 \\
    PC with ``weight''  & 11127 & 247.389 & 3     & 2 \\
    PPC with ``weight'' & 30116 & 511.332 & 2     & 3 \\
    PC with ``cardinality descending''  & 136270 & 6805.031 & 15    & 17 \\
    PPC with ``cardinality descending'' & 90487 & 2558.297 & 10    & 11 \\
    \hline
    \end{tabular}%
    \end{center}
  \label{case2IncSTN}%
  }
\end{table}%

\begin{table}[htbp]
{\relsize{-2}
 \begin{center}
  \caption{Results for learning STNs with mistakes: case 3}
    \begin{tabular}{ | c | c | c | c | c | }
    \hline
          &  {\# of Queries} &  {Time(s)} & \multicolumn{1}{l}{Mistakes} & {Backtracks} \\
          \hline
    BT  & 370901 & 12.257 & 21    & 21 \\
    PC    & 195919 & 4204.597 & 16    & 13 \\
    PC with ``cardinality''  & 12320 & 171.216 & 8     & 1 \\
    PC with ``weight''  & 22223 & 426.241 & 6     & 4 \\
    PC with ``cardinality descending''  & 73722 & 50624.19 & 28    & 22 \\
    \hline
    \end{tabular}%
    \end{center}
  \label{case3IncSTN}%
  }
\end{table}%

\section{Conclusion}
\label{sec:ConclusionAndFutureWork}

Learning QCNs, especially in the case of temporal and spatial information, is of great relevance in many real world applications including scheduling, planning, configuration and Geographic Information Systems (GIS). In this regard, we have proposed a new constraint acquisition algorithm for learning QCNs, through membership queries. Unlike the known constraint learning algorithms, ours has the ability to efficiently learn path consistent networks, which comes handy when it comes to solving the problem. Moreover, a propose a variant of our algorithm to learn QCNs when some of the queries are not correct. This is a significant feature that has been neglected in the literature. In order to achieve these goals, we rely on  path consistency and queries ordering heuristics.   In order to assess the efficiency of our algorithm, in practice, we have conducted several experiments on randomly generated IA and RCC networks, considering several scenarios. The results of these tests are very promising and encouraging.

In the near future, we plan to pursue this work by considering other QCNs such the rectangle algebra \cite{balbiani1999new}, the n-intersections \cite{doi:10.1080/02693799108927841}, as well as those tractable subclasses of IA and RCC8 networks. The advantage when considering these latter networks is that path consistency ensures the consistency of the network. Moreover, in addition to path consistency and ordering heuristics, we can also rely in this case on  conceptual neighbourhood graphs (CNGs) \cite{freksa1992,long2015distributive} and pre-convex relations  (such as ORD-Horn relations)\cite{ligozat1994} to reduce the number of membership queries. We will as well consider temporal constraint networks involving both quantitative and qualitative information \cite{DBLP:journals/apin/MouhoubS12,revesz2009tightened}. Another future direction we will consider, is to extend our algorithm to learning preferences as these often co-exist with constraints in many real world applications \cite{DBLP:journals/scc/MouhoubS08,DBLP:journals/jtaer/SadaouiS16}.

\bibliography{QCNJournal}

\bibliographystyle{plain}



\end{document}